%% file: example_paper.tex
\documentclass{article}

\usepackage{microtype}
\usepackage{graphicx}
\usepackage{subfigure}
\usepackage{booktabs} 
\usepackage[cjk]{kotex}
\usepackage{amssymb}
\usepackage{amsmath}
\usepackage{balance}
\newtheorem{theorem}{Theorem}

\usepackage{natbib}
\usepackage{multibib}

\usepackage{algorithm} 
\usepackage{algorithmic}  
\usepackage[titlenumbered,ruled,algo2e]{algorithm2e}
\allowdisplaybreaks
\newcites{apndx}{References in Supplementary Materials}

\usepackage{hyperref}

\usepackage[accepted]{icml2019}

\icmltitlerunning{HexaGAN: Generative Adversarial Nets for Real World Classification}

\begin{document}

\twocolumn[
\icmltitle{HexaGAN: Generative Adversarial Nets for Real World Classification}



\icmlsetsymbol{equal}{*}

\begin{icmlauthorlist}
\icmlauthor{Uiwon Hwang}{snu}
\icmlauthor{Dahuin Jung}{snu}
\icmlauthor{Sungroh Yoon}{snu,asri}
\end{icmlauthorlist}

\icmlaffiliation{snu}{Electrical and Computer Engineering, Seoul National University, Seoul, Korea}
\icmlaffiliation{asri}{ASRI, INMC, Institute of Engineering Research, Seoul National University, Seoul, Korea}

\icmlcorrespondingauthor{Sungroh Yoon}{sryoon@snu.ac.kr}

\icmlkeywords{Machine Learning, ICML}

\vskip 0.3in
]



\printAffiliationsAndNotice{}  

\begin{abstract}
Most deep learning classification studies assume clean data. However, when dealing with the real world data, we encounter three problems such as 1) missing data, 2) class imbalance, and 3) missing label problems. These problems undermine the performance of a classifier. Various preprocessing techniques have been proposed to mitigate one of these problems, but an algorithm that assumes and resolves all three problems together has not been proposed yet. In this paper, we propose HexaGAN, a generative adversarial network framework that shows promising classification performance for all three problems. We interpret the three problems from a single perspective to solve them jointly. To enable this, the framework consists of six components, which interact with each other. We also devise novel loss functions corresponding to the architecture. The designed loss functions allow us to achieve state-of-the-art imputation performance, with up to a 14\% improvement, and to generate high-quality class-conditional data. We evaluate the classification performance (F1-score) of the proposed method with 20\% missingness and confirm up to a 5\% improvement in comparison with the performance of combinations of state-of-the-art methods.
\end{abstract}

\section{Introduction}
\input{1_introduction}

\section{Generative Adversarial Networks}
\input{2_related_work}

\section{Proposed Method} \label{sec:method}
\input{3_method}

\section{Experiments}
\input{4_experiments}

\section{Conclusion}
\input{5_conclusion}

\balance
\bibliographystyle{icml2019}
\bibliography{example_paper}

\pagebreak
\appendix
\onecolumn

\setcounter{figure}{0}
\setcounter{table}{0}
\setcounter{equation}{0}

\icmltitle{Supplementary Materials \\
HexaGAN: Generative Adversarial Nets for Real World Classification}

\begin{icmlauthorlist}
\icmlauthor{Uiwon Hwang}{snu}
\icmlauthor{Dahuin Jung}{snu}
\icmlauthor{Sungroh Yoon}{snu,asri}
\end{icmlauthorlist}

\icmlaffiliation{snu}{Electrical and Computer Engineering, Seoul National University, Seoul, Korea}
\icmlaffiliation{asri}{ASRI, INMC, Institute of Engineering Research, Seoul National University, Seoul, Korea}

\icmlcorrespondingauthor{Sungroh Yoon}{sryoon@snu.ac.kr}

\icmlkeywords{Machine Learning, ICML}

\vskip 0.3in



\printAffiliationsAndNotice{}  
\setcounter{section}{0}
\input{appendix}

\bibliographystyleapndx{icml2019}
\bibliographyapndx{apndx}

\end{document}

%% file: 1_introduction.tex
As deep learning models have achieved super-human performance in image classification tasks \cite{resnet}, there have been increasing attempts to apply deep learning models to more complicated tasks such as object detection \cite{fastrcnn}, text classification \cite{text}, and disease prediction \cite{uiwon}. However, real world data are often \textit{dirty}, which means that the elements and labels are missing, or there is an imbalance between different classes of data. This prevents a classifier from being fully effective, and thus a preprocessing phase is required. Despite a considerable amount of research, no preprocessing technique has been proposed to address these three problems concurrently. Therefore, we first propose a framework which deals robustly with dirty data.

\begin{figure}[t]
\centering
\includegraphics[width=0.85\linewidth]{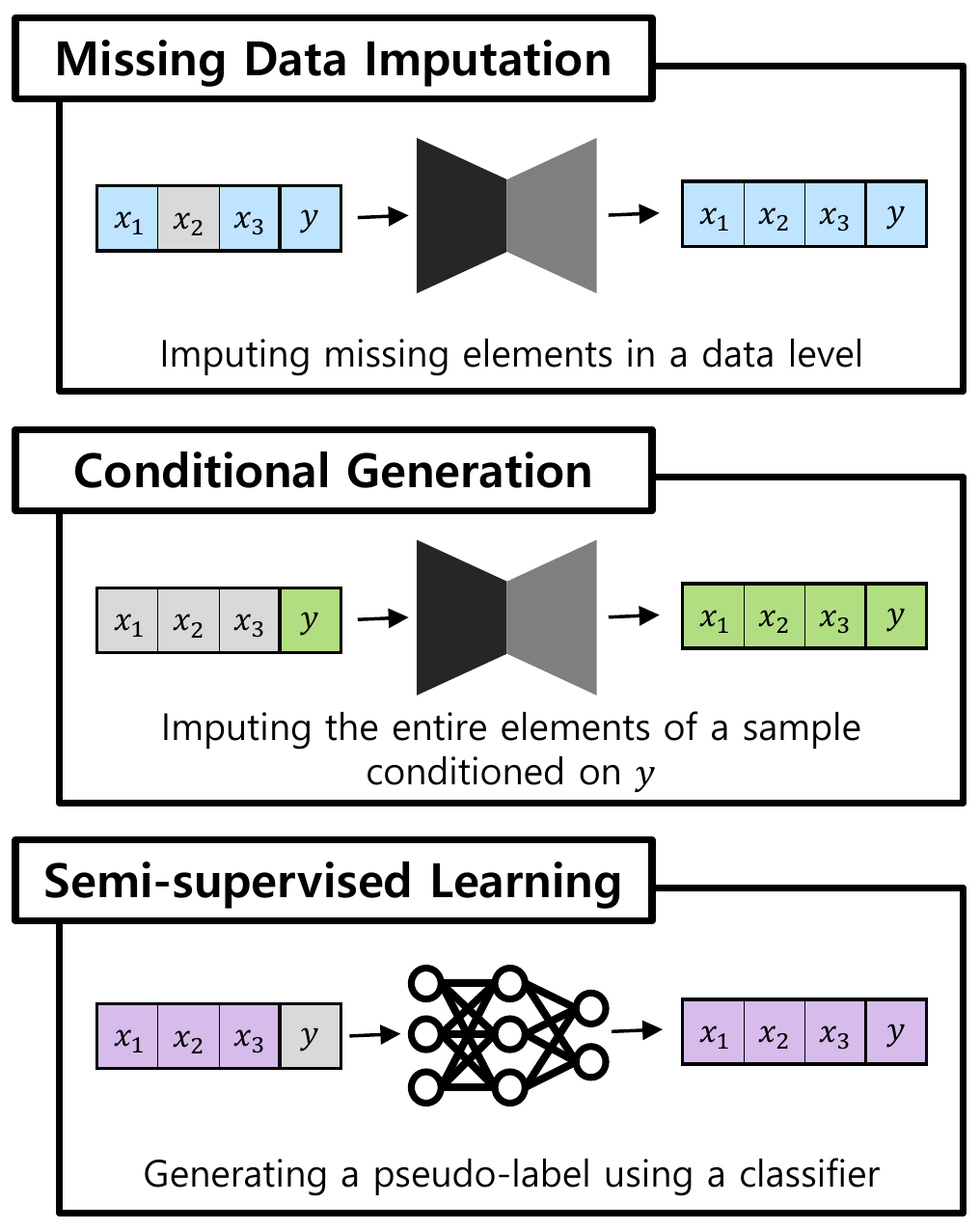} 
\caption{Tasks for the three main problems in real world classification. We define missing data imputation as a task that fills in missing data elements. Conditional generation can be defined as a task that imputes the entire elements in an instance conditioned on a certain class. Semi-supervised learning can be defined as a task that imputes missing labels.}
\label{fig:tasks}
\end{figure}

\begin{figure*}[t]
\centering
\includegraphics[width=1\textwidth]{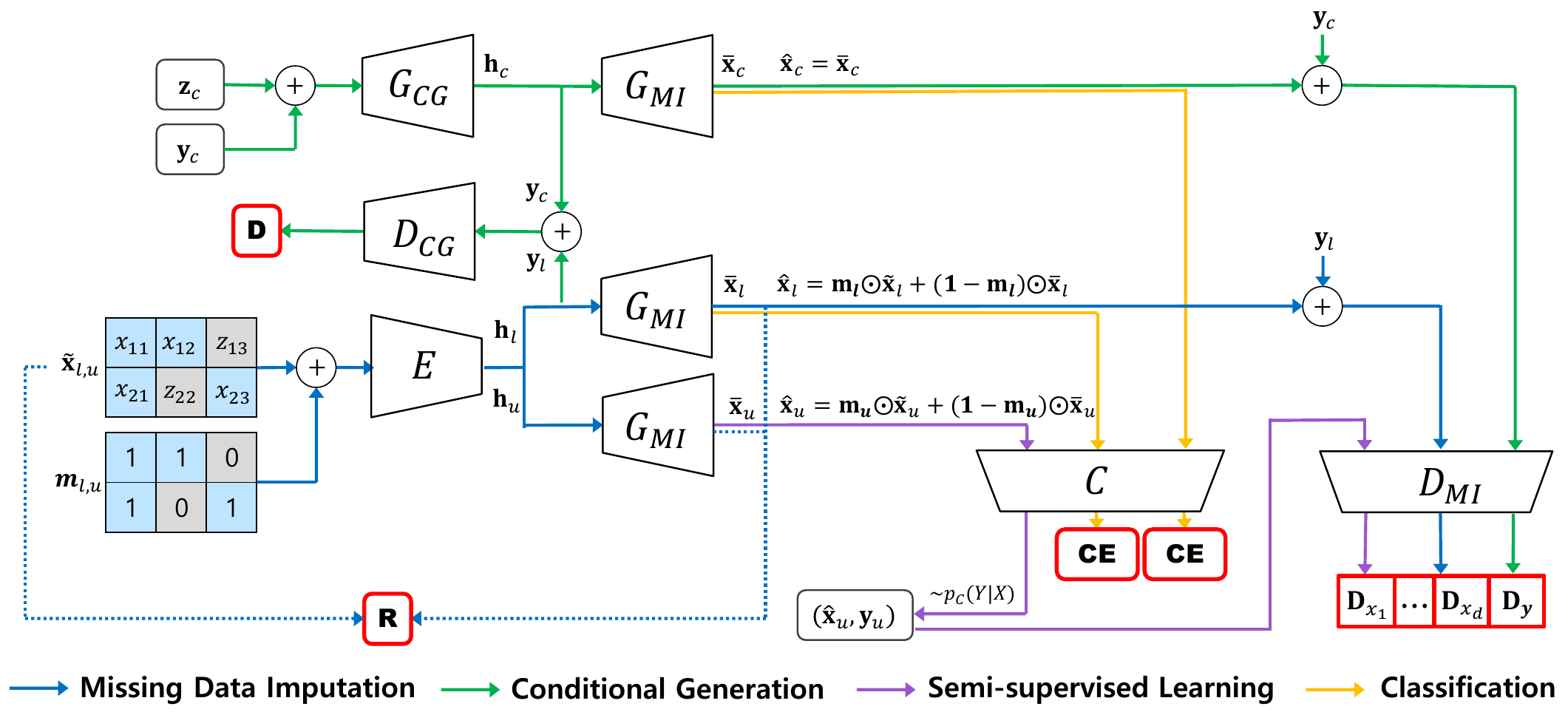} 
\caption{Overview of the HexaGAN model. Subscripts $l$, $u$, and $c$ indicate that a vector is from labeled data, unlabeled data, and class-conditional data respectively. $\mathbf{\tilde{x}}$ denotes a data instance whose missing elements are replaced with noise. $\mathbf{\bar{x}}$ denotes a data generated by $G_{MI}$. $\mathbf{\hat{x}}$ denotes a data instance whose missing elements are filled with the generated values. $\mathbf{y}$ is a class label. Unlike $\mathbf{y}_l$ and $\mathbf{y}_c$, $\mathbf{y}_u$ is produced by $C$. $\mathbf{m}$ is a vector that indicates whether corresponding elements are missing or not. $\mathbf{h}$ is a vector in the hidden space. \textbf{R} is the reconstruction loss. \textbf{D} is the adversarial loss function between $G_{CG}$ and $D_{CG}$. $\mathbf{D}_{x_i}$ represents the element-wise adversarial loss function. $\mathbf{D}_{y}$ represents the adversarial loss function for the label. \textbf{CE} represents the cross-entropy loss.}
\label{fig:overview}
\end{figure*}

The types of data which are typically bedeviled with missing information include the user data employed in recommender systems \cite{factorization}, and in electronic health records \cite{deeppatient} utilizing deep learning based classifiers. \citet{rubin} identifies three main types of missing data: 1) Data are missing completely at random (MCAR). This type of missing data has no pattern which can be correlated with any other variable, whether observed or not. 2) Data are missing at random (MAR). In this case, the pattern of missing data can be correlated with one or more observed variables. 3) Data are missing (but) not at random (MNAR). The pattern of this type of missing data can be related to both observed and unobserved variables. In this paper, we are concerned with MCAR data. The replacement of missing information within data is called \textit{imputation} \cite{imputation}. Imputation techniques include matrix completion \cite{softimpute}, k-nearest neighbors \cite{knn}, multivariate imputation by chained equations (MICE) \cite{mice}, denoising autoencoders \cite{dae}, and methods based on generative adversarial networks (GAN) \cite{gain, vigan}. Poor or inappropriate imputation can mislead deep learning based techniques into learning the wrong data distribution.
Many real world datasets such as those related to anomaly detection \cite{anomaly} and disease prediction \cite{disease_imbalance} involve poorly balanced classes. The class imbalance problem can be overcome by techniques such as the synthetic minority oversampling technique (SMOTE) \cite{smote} and adaptive synthetic (ADASYN) sampling \cite{adasyn}. However, oversampling from the entire data distribution requires a large amount of memory. Cost sensitive loss \cite{cs} is also used to solve the class imbalance problem by differentiating cost weights to each class. However, cost sensitive loss tends to overfit to the minority classes \cite{imbalance}. We overcome the class imbalance problem by training a deep generative model to follow the true data distribution, and then generate samples of minority classes for each batch. This requires conditional generation, which we regard as imputation, as shown in Figure \ref{fig:tasks}, in which entire elements are imputed according to the appropriate class label.

In deep learning, the amount of labeled training data has a significant impact on the performance. Insufficiency of labeled data is referred to as the missing label problem. It is encountered in real world applications such as natural language models \cite{nlp_semi} or healthcare systems \cite{healthcare_semi}, where the cost of labeling is expensive. Related researchers have proposed semi-supervised methods by which to leverage unlabeled data. Semi-supervised learning is designed to make the best use of unlabeled data, using regularization and generative approaches. The regularization approach adds a regularization loss term, which is designed on the assumption that adjacent data points or the same architectural data points are likely to have the same label \cite{label_prop, phi, entmin, vat, mean_teacher}. Unlike the regularization approach, the generative approach enhances the performance of a classifier by utilizing raw unlabeled data in training the generative model \cite{kingma, semi_vae, sslgan, catgan, badgan}.

 As depicted in Figure \ref{fig:tasks}, we define the missing data, class imbalance, and missing label problems in terms of imputation. Based on insight concerning the imputation, we find out that networks used for imputation can play multiple roles. Moreover, solving the three data problems simultaneously is more effective than solving them in a cascading form. In this paper, we propose a GAN framework consisting of six components to solve the three problems in real world classifications. We derive a new objective function for the imputation of missing data, and demonstrate that it performs better than the existing state-of-the-art imputation methods. We define conditional generation from the perspective of conditional imputation, and confirm that the proposed method works successfully by designing the imputation model to be a part of the framework. In order to deal with the missing label problem, we use semi-supervised learning, in which a classifier generates a synthetic class label for unlabeled data and a discriminator distinguishes fake from real labels.

In summary, our contributions are as follows:
\begin{itemize}
    \item 
    To the best of our knowledge, this is one of the first studies that defines the three problems (missing data, class imbalance, and missing label) in terms of imputation. Then, we propose HexaGAN to encourage thorough imputation of data with these three problems.
    \item 
    To implement real world datasets into existing classifiers, we must apply suitable preprocessing techniques to the datasets. However, our framework is simple to use and works automatically when the absence of data elements and labels is indicated ($\mathbf{m}$ and $m_y$, See Section \ref{sec:method}).
    \item 
    We devise a combination of six components and the corresponding cost functions. More specifically, we propose a novel adversarial loss function and gradient penalty for element-wise imputation, confirming that our imputation performance produces stable, state-of-the-art results.
    \item 
    In real world classification, the proposed method significantly outperforms cascading combinations of the existing state-of-the-art methods. As a result, we demonstrate that the components of our framework interplay to solve the problems effectively.
    
\end{itemize}

%% file: 2_related_work.tex
Generative models, which include GANs \cite{gan}, are capable of generating high-quality synthetic data for many applications. Although GANs are the most advanced than other techniques, model training can be unstable. Many studies have tried to stabilize GANs. Among them, \citet{wgan} proposed the Wasserstein GAN (WGAN), which has a smoother gradient by introducing the Wasserstein (Earth Mover) distance. Several gradient penalties have also been proposed \cite{wgan-gp,zc} to make WGAN training more stable. In this paper, we modify the WGAN loss and zero-centered gradient penalty for missing data imputation. Experimentally, we show that the proposed method has more stable and better imputation performance than the existing vanilla GAN loss-based model.

GAIN \cite{gain} is the first method to use a GAN for imputing MCAR data. The typical discriminator predicts whether each \textit{instance} is real or fake. However, this task is difficult if all instances have missing data. Instead, GAIN labels each \textit{element} of an instance as missing or not, so that the discriminator can discriminate between real and fake elements. Our imputation method shares some similarity with GAIN because both methods label elements as real or fake. The imputation performance of GAIN measured by our own implementation with the specific dataset is lower than that of the autoencoder, and the learning curve appears to be unstable. However, HexaGAN provides a stable imputation performance, and usability by including class-conditional generation to address the class imbalance problem, and through the use of semi-supervised learning.

TripleGAN \cite{triplegan} is a GAN for semi-supervised learning in which a classifier, a generator, and a discriminator interact. The classifier creates pseudo-labels for unlabeled data, and image-label pairs are then passed to the discriminator. The classifier and discriminator are trained competitively. In this paper, we adopt the pseudo-labeling technique of TripleGAN to allow HexaGAN to perform semi-supervised learning.

%% file: 3_method.tex

The HexaGAN framework is comprised of six components, as illustrated in Figure \ref{fig:overview}:
\begin{itemize}
    \item $E$: the encoder, that transfers both labeled and unlabeled instances into the hidden space.
    \item $G_{MI}$: a generator that imputes missing data.
    \item $D_{MI}$: a discriminator for missing imputation, that distinguishes between missing and non-missing elements and labels.
    \item $G_{CG}$: a generator that creates conditional hidden vectors $\mathbf{h}_c$.
    \item $D_{CG}$: a discriminator for conditional generation, that determines whether a hidden vector is from the dataset or has been created by $G_{CG}$.
    \item $C$: the classifier, that estimates class labels. This also works as the label generator.
\end{itemize}

HexaGAN operates on datasets containing instances $\mathbf{x}^1, ..., \mathbf{x}^n \in \mathbb{R}^d$, where $n$ is the number of instances and $d$ is the number of elements in an instance. The $i$-th element in a single instance $x_i^j$ is a scalar, and some of these elements may be missing. The first $n_l$ instances are labeled data, and the remaining $n-n_l$ instances are unlabeled data. There are class labels $\mathbf{y}^1, ..., \mathbf{y}^{n_l} \in \mathbb{R}^{n_c}$ corresponding to each instance, where $n_c$ is the number of classes. Boolean vectors $\mathbf{m}^1, ..., \mathbf{m}^n \in \mathbb{R}^{d}$ indicate whether each element in an instance is missing or not. If $m_i^j$ (the $i$-th element of a vector $\mathbf{m}^j$) is $0$, $x_i^j$ is missing. The boolean $m_y \in \mathbb{R}$ indicates whether an instance has a label or not. If $m_y$ is $0$, the label is missing. Thus, labeled instances exist as a set of $D_l=\{(\mathbf{x}^j, \mathbf{y}^j, \mathbf{m}^j, m_y^j=1)\}_{j=1}^{n_l}$, and unlabeled instances exist as a set of $D_u=\{(\mathbf{x}^j, \mathbf{m}^j, m_y^j=0)\}_{j=n_l+1}^{n}$.

\subsection{Missing data imputation}
Missing data imputation aims to fill in missing elements using the distribution of data represented by the generative model. In HexaGAN, missing data imputation is performed by $E$, $G_{MI}$, and $D_{MI}$. An instance received by $D_{MI}$ is not labeled as real or fake, but each \textit{element} is labeled as real (non-missing) or fake (missing).

From now on, we omit the superscript for a clearer explanation (i.e., $x_i^j$=$x_i$). First, we make a noise vector $\mathbf{z} \in \mathbb{R}^d$ with the same dimension as an input instance $\mathbf{x} \in (\mathbf{x}_l \cup \mathbf{x}_u)$ by sampling from a uniform distribution $U(0, 1)$. We replace the missing elements in the instance with elements of $\mathbf{z}$ to generate $\tilde{\mathbf{x}}$:
\begin{gather}
\tilde{\mathbf{x}} = \mathbf{m} \odot \mathbf{x} + (\mathbf{1}-\mathbf{m}) \odot \mathbf{z}
\end{gather}
where $\odot$ is element-wise multiplication. The objective of our framework is to sample the patterns stored in the model that are the most suitable replacements for the missing data (i.e., to generate samples which follow $p(\mathbf{x}|\tilde{\mathbf{x}},\mathbf{m})$). Then, $\tilde{\mathbf{x}}$ is concatenated with $\mathbf{m}$, and from the pair $(\tilde{\mathbf{x}}, \textbf{m})$, the encoder $E$ generates a hidden variable $\mathbf{h} = E(\tilde{\mathbf{x}}, \textbf{m})$ in the hidden space, which has the dimension $d_\mathrm{h}$.

The $G_{MI}$ receives $\mathbf{h}$ and generates $\bar{\mathbf{x}} = G_{MI}(\mathbf{h})$. The missing elements in the input instance are imputed with the generated values, resulting in $\hat{\mathbf{x}}$ as follows:
\begin{gather}
    \hat{\mathbf{x}} = \mathbf{m} \odot \mathbf{x} + (\mathbf{1}-\mathbf{m}) \odot \bar{\mathbf{x}}
\end{gather}
The $D_{MI}$ now determines whether each element of the pair ($\hat{\mathbf{x}}, \mathbf{y}$) is real or fake. The label for $\mathbf{\hat{x}}$ is $\mathbf{m}$. The $D_{MI}$ calculates the adversarial losses by determining whether the missingness is correctly predicted for each element, which is then used to train $E$, $G_{MI}$, and $D_{MI}$. The adversarial loss $\mathcal{L}_{G_{MI}}$ which is used to train $E$ and $G_{MI}$, and $\mathcal{L}_{D_{MI}}$ which is used to train $D_{MI}$ can be expressed as follows:
\begin{align}
    \mathcal{L}_{G_{MI}} = -\sum_{i=1}^{d} &\mathbb{E}_{\hat{\mathbf{x}}, \mathbf{y}, \mathbf{m}}\left[\left(1-m_i\right) \cdot D_{MI}(\hat{\mathbf{x}},\mathbf{y})_i\right] \\
    \mathcal{L}_{D_{MI}} = \sum_{i=1}^{d} &\mathbb{E}_{\hat{\mathbf{x}}, \mathbf{y}, \mathbf{m}}\left[\left(1-m_i\right) \cdot D_{MI}(\hat{\mathbf{x}}, \mathbf{y})_i\right] \label{eq:L_D_MI} \\\notag 
    &- \mathbb{E}_{\hat{\mathbf{x}}, \mathbf{y}, \mathbf{m}}\left[m_i \cdot D_{MI}(\hat{\mathbf{x}}, \mathbf{y})_i\right] 
\end{align}
where $D_{MI}(\cdot)_i$ is the $i$-th output element of $D_{MI}$. The following theorem confirms that the proposed adversarial loss functions make the generator distribution converge to the desired data distribution.

\begin{algorithm}[tb]
    \SetKwInOut{Input}{input}
    \SetKwInOut{Output}{output}
    \Input{$\textbf{x}$ - data with missing values sampled from $D_l$ and $D_u$; 
    \newline $\textbf{m}$ - vector indicating whether elements are missing; 
    \newline $\textbf{z}$ - noise vector sampled from $U(0, 1)$}
    \Output{$\hat{\textbf{x}}$ - imputed data}
    \caption{Missing data imputation}
    \label{alg:imputation}
\begin{algorithmic}
   \REPEAT
   \STATE Sample a batch of pairs $(\textbf{x}, \textbf{m}, \textbf{z})$
   \STATE $\tilde{\textbf{x}} \leftarrow \textbf{m} \odot \textbf{x} + (\textbf{1}-\textbf{m}) \odot \textbf{z}$
   \STATE $\textbf{h} \leftarrow E(\tilde{\textbf{x}}, \textbf{m})$
   \STATE $\bar{\textbf{x}} \leftarrow G_{MI}(\textbf{h})$
   \STATE $\hat{\textbf{x}} \leftarrow \textbf{m} \odot \textbf{x} + (\textbf{1}-\textbf{m}) \odot \bar{\textbf{x}}$
   \STATE Update $D_{MI}$ using stochastic gradient descent (SGD)
   \STATE $\nabla_{D_{MI}} \mathcal{L}_{D_{MI}} + \lambda_1 \mathcal{L}_{\mathrm{GP}_{MI}}$
   \STATE Update $E$ and $G_{MI}$ using SGD 
   \STATE $\nabla_{E} \mathcal{L}_{G_{MI}} + \alpha_1 \mathcal{L}_{\mathrm{recon}}$ 
   \STATE $\nabla_{G_{MI}} \mathcal{L}_{G_{MI}} + \alpha_1 \mathcal{L}_{\mathrm{recon}}$
   \UNTIL{training loss is converged}
\end{algorithmic}
\end{algorithm}

\begin{theorem} \label{th:optimality}
    A generator distribution $p(\mathbf{x}|\mathbf{m}_i=0)$ is a global optimum for the min-max game of $G_{MI}$ and $D_{MI}$, if and only if $p(\mathbf{x}|\mathbf{m}_i=1)=p(\mathbf{x}|\mathbf{m}_i=0)$ for all $\mathbf{x}\in\mathbb{R}^d$, except possibly on a set of zero Lebesgue measure.
\end{theorem}
Proof of Theorem \ref{th:optimality} is provided in Supplementary Materials.

Moreover, we add a reconstruction loss to the loss function of $E$ and $G_{MI}$ to exploit the information of non-missing elements, as follows:
\begin{align}
    \mathcal{L}_{\mathrm{recon}} &= \mathbb{E}_{\mathbf{\bar{x}}|\mathbf{x},\mathbf{m}}\left[\sum_{i=1}^d m_i (x_i - \bar{x}_i)^2\right]
\end{align}
For more stable GAN training, we modify a simplified version of the zero-centered gradient penalty proposed by \citet{zc} in an element-wise manner, and add the gradient penalty to the loss function of $D_{MI}$. The modified regularizer penalizes the gradients of each output unit of the $D_{MI}$ on $p_\mathcal{D}(x_i)$:
\begin{align}
    \mathcal{L}_{\mathrm{GP}_{MI}} = \sum_{i=1}^d \mathbb{E}_{p_\mathcal{D}(x_i)}\left[||\nabla_{\hat{\mathbf{x}}} D_{MI}(\hat{\mathbf{x}})_i||_2^2\right]
\end{align}
We define $\hat{\mathbf{x}}$ in $p_\mathcal{D}(x_i)$ as data with $m_i$ is $1$ (i.e., $p_\mathcal{D}(x_i)=\lbrace \hat{\mathbf{x}}^j|m_i^j=1 \rbrace$). In other words, as suggested by \citet{zc}, we penalize $D_{MI}$ only for data wherein the $i$-th element is not missing (real) in a batch. This helps balance an adversarial relationship between the generator and discriminator by forcing the discriminator closer to Nash Equilibrium. 

Therefore, missing data imputation and model training are performed as described in Algorithm \ref{alg:imputation}. We used 10 for both hyperparameters $\lambda_1$ and $\alpha_1$ in our experiments.

\subsection{Conditional generation}
We define conditional generation for the class imbalance problem as the imputation of entire data elements on a given class label (i.e., generating $(x_1, ..., x_d)$ following $p(\mathbf{x}|\mathbf{y})$). Since we have $G_{MI}$, which is a generator for imputation, we can oversample data instances by feeding synthetic $\mathbf{h}$ into $G_{MI}$. Therefore, we introduce $G_{CG}$ to generate a hidden variable $\mathbf{h}_c$ corresponding to the target class label $\mathbf{y}_c$, i.e., we sample $\mathbf{h}_c \sim p_{G_{CG}}(\mathbf{h}|\mathbf{y})$. We also introduce $D_{CG}$ to distinguish pairs of generated hidden variables and target class labels $(\mathbf{h}_c, \mathbf{y}_c)$ (fake) from pairs of hidden variables for labeled data and corresponding class labels $(\mathbf{h}_l, \mathbf{y}_l)$ (real). $G_{CG}$ and $D_{CG}$ are trained with WGAN loss and zero-centered gradient penalty on $\mathbf{h}_l$ as follows:
\begin{align}
    \mathcal{L}_{G_{CG}} = & - \mathbb{E}_{\mathbf{h}_c \sim p_{G_{CG}}(\mathbf{h}_c|\mathbf{y}_c)}[D_{CG}(\mathbf{h}_c, \mathbf{y}_c)] \\
    \mathcal{L}_{D_{CG}} = & \mathbb{E}_{\mathbf{h}_c \sim p_{G_{CG}}(\mathbf{h}_c|\mathbf{y}_c)}[D_{CG}(\mathbf{h}_c, \mathbf{y}_c)] \\\notag
    &- \mathbb{E}_{\mathbf{h}_l \sim p_{E}(\mathbf{h}_l|x_l)}[D_{CG}(\mathbf{h}_l, \mathbf{y}_l)] \\
    \mathcal{L}_{\mathrm{GP}_{CG}} = & \mathbb{E}_{\mathbf{h}_l \sim p_{E}(\mathbf{h}_l|x_l)}\left[||\nabla_{\mathbf{h}_l} D_{CG}(\mathbf{h}_l, \mathbf{y}_l)||_2^2\right]
\end{align}

$G_{MI}$ maps generated $\mathbf{h}_c$ into a realistic $\hat{\mathbf{x}}_c$. Because $\mathcal{L}_{G_{CG}}$ is not enough to stably generate $\mathbf{h}_c$, we add the loss of $G_{MI}$ from $\hat{\mathbf{x}}_c$. Since we defined conditional generation as imputation of all the elements, $G_{CG}$ and $D_{MI}$ are related adversarially. The label of $(\mathbf{\hat{x}}_c, \mathbf{y}_c)$ for $D_{MI}$ is a ($d+1$)-dimensional zero vector.

In addition, the cross-entropy of $(\mathbf{\hat{x}}_c, \mathbf{y}_c)$ calculated from the prediction of $C$ is also added to the loss function of $G_{CG}$ to stably generate the data that is conditioned on the target class as follows:
\begin{align}
\mathcal{L}_{\mathrm{CE}}(\mathbf{\hat{x}}_c, \mathbf{y}_c) = -\mathbb{E}_{\mathbf{\hat{x}}_c|\mathbf{y}_c} \left[\sum_{k=1}^{n_c} \mathbf{y}_{c_k} \log(C(\mathbf{\hat{x}}_c)_k)\right]
\end{align} 
where $C(\cdot)_k$ is the softmax output for the $k$-th class.
Thus, $D_{CG}$ and $G_{CG}$ are trained according to:
\begin{align}
\min_{D_{CG}} & \mathcal{L}_{D_{CG}} + \lambda_2 \mathcal{L}_{\mathrm{GP}_{CG}} \\
\min_{G_{CG}} & \mathcal{L}_{G_{CG}} + \alpha_2 \mathcal{L}_{G_{MI}} + \alpha_3 \mathcal{L}_{\mathrm{CE}}(\mathbf{\hat{x}}_c, y_c)
\end{align} 
where $\lambda_2$, $\alpha_2$, and $\alpha_3$ denote hyperparameters, and we set $\lambda_2$ to 10, $\alpha_2$ to 1, and $\alpha_3$ to 0.01 in our experiments. Since the distribution of $\mathbf{h}_l$ moves according to the training of $E$, we set the number of update iterations of $D_ {CG}$ and $G_{CG}$ per an update of $E$ to 10, so that $\mathbf{h}_c$ follows the distribution of $\mathbf{h}_l$ well.

\subsection{Semi-supervised classification}
\subsubsection{Pseudo-labeling}
We define semi-supervised learning as imputing missing labels by the pseudo-labeling technique, TripleGAN \cite{triplegan}. Semi-supervised learning is achieved by the interaction of $C$ and $D_{MI}$. $C$ generates a pseudo-label $\mathbf{y}_u$ of an unlabeled instance $\mathbf{\hat{x}}_u$, i.e., $\mathbf{y}_u$ is sampled from the classifier distribution $p_C(\mathbf{y}|\mathbf{x})$. Then, the data-label pair ($\mathbf{\hat{x}}_u$, $\mathbf{y}_u$) enters $D_{MI}$. The last element of the $D_{MI}$ output, $D_{MI}(\cdot)_{d+1}$, determines whether the label is real or fake. The label for pseudo-labeling is $m_y$. $C$ and $D_{MI}$ are trained according to the following loss functions:
\begin{align}
    \mathcal{L}_C = - &\mathbb{E}_{\mathbf{y}_u|\mathbf{\hat{x}}_u \sim p_C} \left[ D_{MI}(\mathbf{\hat{x}}_u, \mathbf{y}_u)_{d+1} \right] \\
    \mathcal{L}_{D_{MI}}^{d+1} = &\mathbb{E}_{\mathbf{y}_u|\mathbf{\hat{x}}_u \sim p_C} \left[ D_{MI}(\mathbf{\hat{x}}_u, \mathbf{y}_u)_{d+1} \right] \\\notag
    &- \mathbb{E}_{\mathbf{y}|\mathbf{\hat{x}} \sim p_{data}} \left[ D_{MI}(\mathbf{\hat{x}}, \mathbf{y})_{d+1} \right]
\end{align} 
where $p_{data}$ denotes the data distribution of y conditioned on $\mathbf{\hat{x}}$. $\mathcal{L}_{D_{MI}}^{d+1}$ is added to the loss of $D_{MI}$, so that $i$ in Equation \ref{eq:L_D_MI} expands from $d$ to $d+1$. If $G_{MI}$ learns the true data distribution, then we can postulate that $p_{data}$ follows the true conditional distribution. We should note that the adversarial loss is identical to the loss function of WGAN between $C$ and $D_{MI}$. Therefore, $C$ plays a role as a label generator, and $D_{MI}(\cdot)_{d+1}$ acts as a label discriminator.

Through adversarial learning, we expect that the adversarial loss enhances the performance of $C$. It can be shown that $C$ minimizing the adversarial loss $\mathcal{L}_C$ is equivalent to optimizing the output distribution matching (ODM) cost \cite{odm}. 

\begin{theorem} \label{th:odm}
Optimizing the adversarial losses for $C$ and $D_{MI}(\cdot)_{d+1}$ imposes an unsupervised constraint on $C$. Then, the adversarial losses for semi-supervised learning in HexaGAN satisfy the definition of the ODM cost.
\end{theorem}
Proof of Theorem \ref{th:odm} is provided in Supplementary Materials.

According to the properties of the ODM cost, the global optimum of supervised learning is also a global optimum of semi-supervised learning. Therefore, intuitively, $\mathcal{L}_C$ and $\mathcal{L}_{D_{MI}}^{d+1}$ serve as guides for finding the optimum point of the supervised loss.

\subsubsection{Classification of HexaGAN}
In order to train $C$, the two models $E$ and $G_{MI}$ impute the missing values of data instances $\mathbf{\hat{x}}_l$. $G_{CG}$ produces hidden vectors $\mathbf{h}_c$ conditioned on the minority classes so that the number of data in the minority classes in each batch is equal to the number of data instances in the majority class of each batch, and $G_{MI}$ generates class-conditional data $\mathbf{\hat{x}}_c$. Then, the cross-entropy between $\mathbf{\hat{x}}_{l,c} \in (\mathbf{\hat{x}}_l \cup \mathbf{\hat{x}}_c)$ and $\mathbf{y}_{l,c} \in (\mathbf{y}_l \cup \mathbf{y}_c)$ is calculated to train $C$. Unlabeled data $\mathbf{\hat{x}}_u$ is used to optimize $L_C$, the loss for pseudo-labeling, thereby training a more robust classifier.

Therefore, $C$ is trained according to:
\begin{align}
\min_{C} \mathcal{L}_{\mathrm{CE}}(\mathbf{\hat{x}}_{l,c}, \mathbf{y}_{l,c}) + \alpha_4 \mathcal{L}_C
\end{align}
where we used 0.1 for $\alpha_4$ in our experiments. The entire training procedure of HexaGAN is presented in Supplementary Materials.

%% file: 4_experiments.tex
\begin{figure}[t]
\centering
\includegraphics[width=0.9\linewidth]{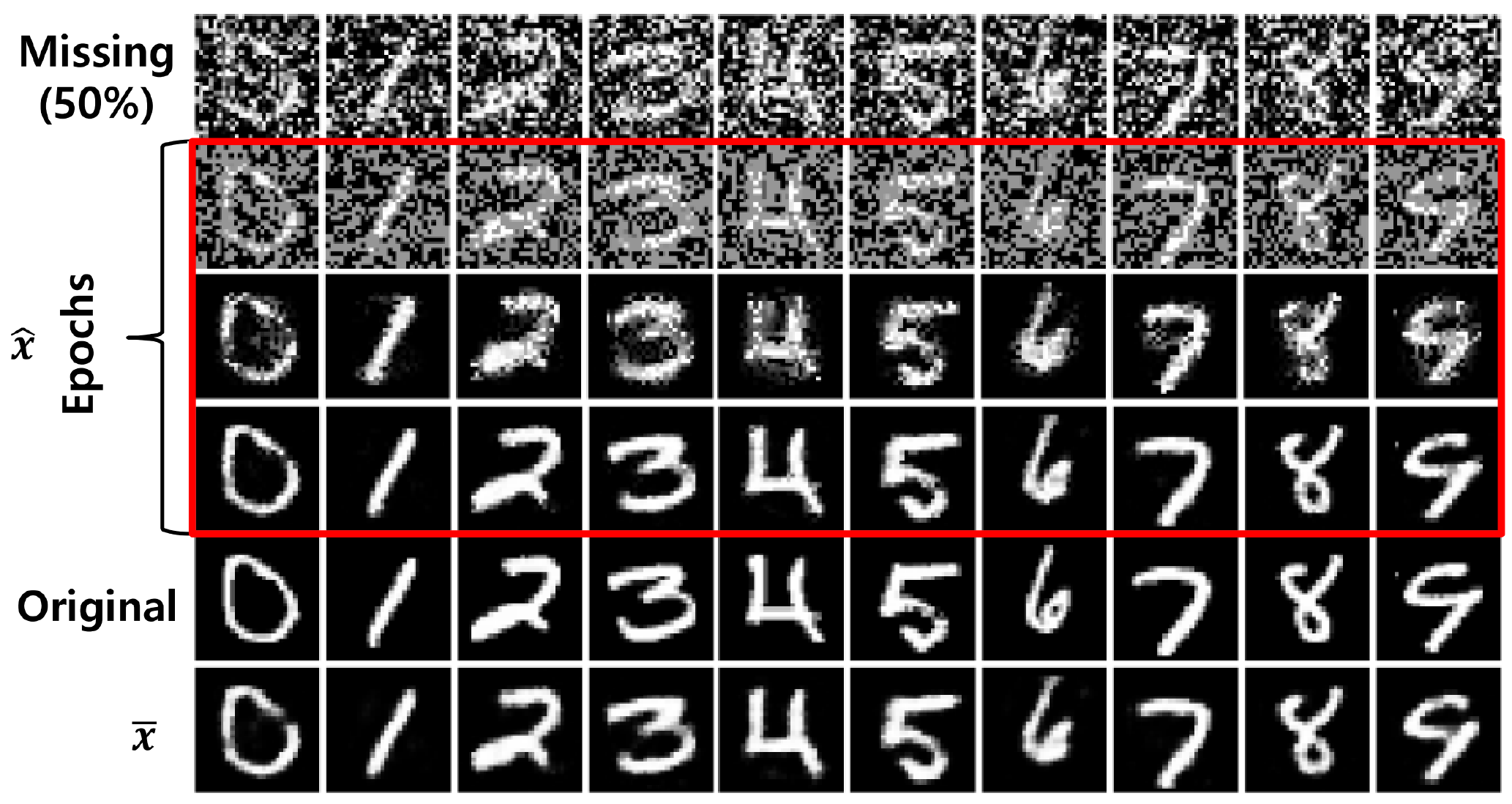} 
\caption{Imputation results with the MNIST dataset. 1st row: MNIST images with 50\% missing randomly as inputs of HexaGAN. 2nd$\sim$4th rows (red box): images imputed by HexaGAN ($\mathbf{\hat{x}}$) at 1, 10, and 100 epochs. 5th row: original images (no missing element). 6th row: images generated by $G_{MI}$ for imputation ($\mathbf{\bar{x}}$).}
\label{fig:imputation}
\end{figure}

Here, we present the performance of the proposed method. We used datasets from the UCI machine learning repository \cite{uci}, including real world datasets (breast, credit, wine) and a synthetic dataset (madelon). Detailed descriptions are presented in Supplementary Materials. We also used a handwritten digit dataset (MNIST). First, we show the imputation performance of HexaGAN. Then, we show the quality of conditional generation using our framework. Finally, we present the classification performance of our proposed model, assuming the problems in real world classification.

We basically assume 20\% missingness (MCAR) in the elements and labels of the UCI dataset and 50\% in the elements of the MNIST dataset to cause missing data and missing label problems. Every element was scaled to a range of [0,1]. We repeated each experiment 10 times and used 5-fold cross validation. As the performance metric, we calculated the root mean square error (RMSE) for missing data imputation and the F1-score for classification. We analyzed the learning curve and found that the modified zero-centered gradient and RMSprop promote stability in HexaGAN. The details are described in Supplementary Materials. The architecture of HexaGAN can also be found in Supplementary Materials.

\begin{table}[t!]
\caption{Performance comparison with other imputation methods (RMSE)}
\label{tab:imputation}
{\resizebox{1\columnwidth}{!}
{\begin{tabular}{c|c|c|c|c|c}

\hline
	\toprule
   Method & Breast & Credit & Wine & Madelon & MNIST\\
   \midrule
    Zeros & 0.2699 & 0.2283 & 0.4213 & 0.5156 & 0.3319 \\
    Matrix & 0.0976 & 0.1277 & 0.1772 & 0.1456 & 0.2540 \\
    K-NN & 0.0872 & 0.1128 & 0.1695 & 0.1530 & 0.2267 \\
    MICE & 0.0842 & 0.1073 & 0.1708 & 0.1479 & 0.2576 \\
    Autoencoder & 0.0875 & 0.1073 & 0.1481 & 0.1426 & 0.1506 \\
    GAIN & 0.0878 & 0.1059 & 0.1406 & 0.1426 & 0.1481 \\
    HexaGAN & \textbf{0.0769} & \textbf{0.1022} & \textbf{0.1372} & \textbf{0.1418} & \textbf{0.1452} \\
    \bottomrule
\end{tabular}}
}
\end{table}

\begin{figure*}[t]
\centering
\includegraphics[width=0.9\linewidth]{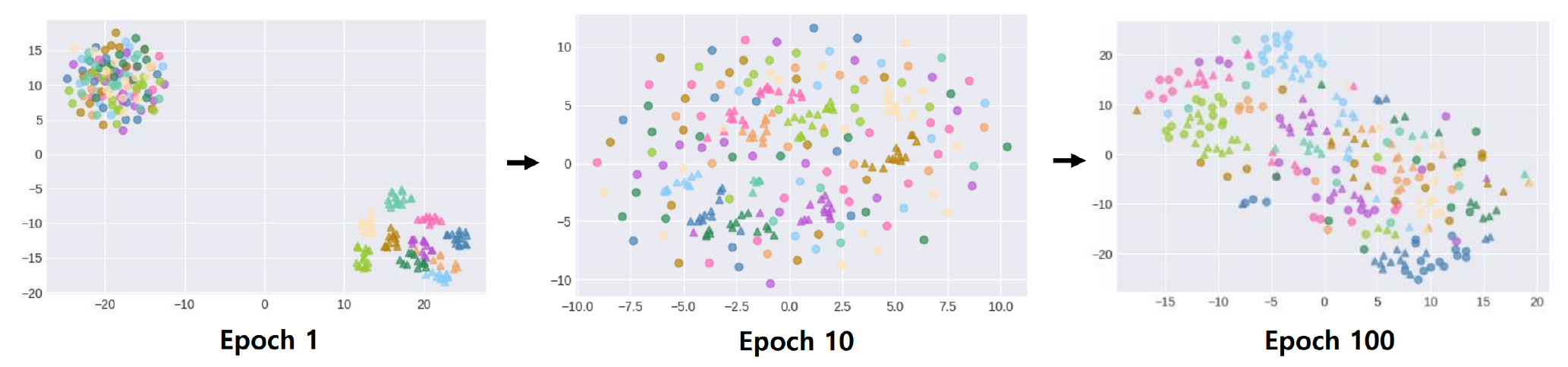} 
\caption{tSNE analysis with the MNIST dataset at 1, 10, and 100 epochs. The circles stand for $\mathbf{h}_l$ (hidden vectors from $E$). The triangles denotes $\mathbf{h}_c$ (hidden vectors from $G_{CG}$). Different colors represent different class labels.}
\label{fig:tsne}
\end{figure*}

\begin{figure}[t]
\centering
\includegraphics[width=0.9\linewidth]{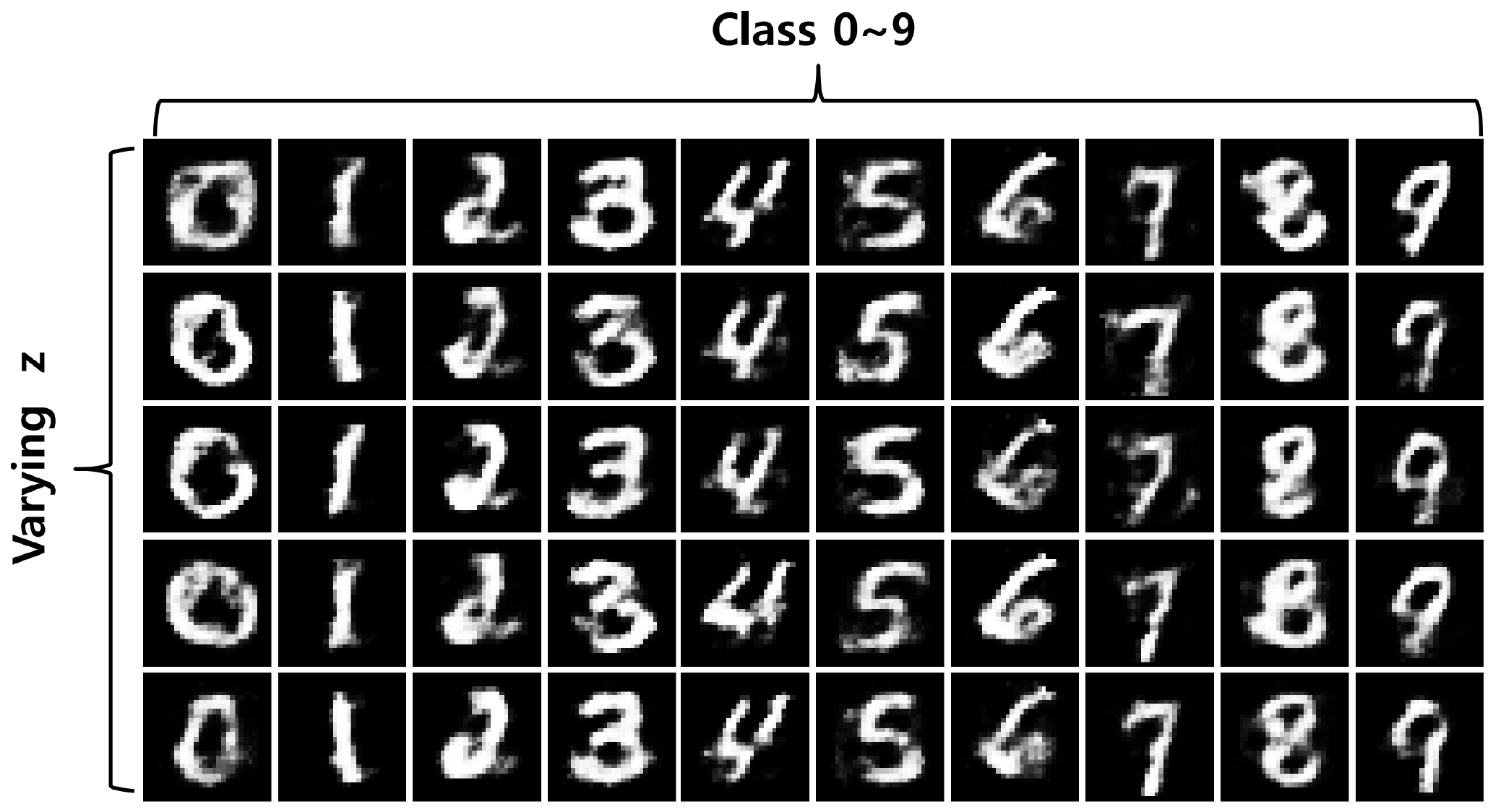} 
\caption{Class-conditional generation results with the MNIST dataset. Each row visualizes generated images conditioned on 0$\sim$9. Each column shows images generated by all different $\mathbf{z}$s.}
\label{fig:cg}
\end{figure}

\subsection{Imputation performance}
\subsubsection{Comparison with real world datasets}
We used UCI datasets and the MNIST dataset to evaluate the imputation performance. Table \ref{tab:imputation} shows the imputation performance of zero imputation, matrix completion, k-nearest neighbors, MICE, autoencoder, GAIN, and HexaGAN. In our experiments, we observed that HexaGAN outperforms the state-of-the-art methods on all datasets (up to a 14\% improvement). Two deep generative models, GAIN and HexaGAN, use both the reconstruction loss and the adversarial loss. GAIN shows the same or lower performance than the autoencoder on certain datasets, whereas HexaGAN consistently outperforms the autoencoder on all datasets. This shows that the novel adversarial loss boosts the imputation performance.

\subsubsection{Qualitative analysis} \label{sec:mnist_imputation}
Figure \ref{fig:imputation} visualizes the imputation performance with the MNIST dataset. Since MNIST is an image dataset, we designed HexaGAN with convolutional and deconvolutional neural networks. The first row of Figure \ref{fig:imputation} shows MNIST data with 50\% missing as the input for HexaGAN. The next three rows show $\mathbf{\hat{x}}$ after 1, 10, and 100 epochs, and it can be seen that higher quality imputed data are generated as the number of epochs increases. The next row presents the original data with no missing value, and the last row shows $\mathbf{\bar{x}}$ generated by $G_{MI}$. This suggests that the proposed method imputes missing values with very high-quality data. The RMSE value using the convolutional architecture is 0.0914.

\subsection{Conditional generation performance}
\subsubsection{TSNE analysis}
We used tSNE \cite{tsne} to analyze $\mathbf{h}_l$ generated by $E$ and $\mathbf{h}_c$ generated by $G_{CG}$. Figure \ref{fig:tsne} shows the changes of $\mathbf{h}_l$ (circle) and $\mathbf{h}_c$ (triangle) according to the iteration. Each color stands for a class label. At epoch 1, $\mathbf{h}_l$ and $\mathbf{h}_c$ have very different distributions, and form respective clusters. At epoch 10, the cluster of $\mathbf{h}_c$ is overlapped by the cluster of $\mathbf{h}_l$. At epoch 100, $E$ learns the manifold of the hidden representation, so that $\mathbf{h}_l$ is gathered by class and $\mathbf{h}_c$ follows the distribution of $\mathbf{h}_l$ well. That is, $G_{CG}$ creates a high-quality $\mathbf{h}_c$ that is conditioned on a class label. The complete version of the tSNE analysis is given in Supplementary Materials.

\subsubsection{Qualitative analysis}
To evaluate the performance of conditional generation, we used the same architecture as in Section \ref{sec:mnist_imputation} and generated synthetic MNIST images conditioned on 10 class labels. Figure \ref{fig:cg} presents the generated MNIST images. Each row shows the results of conditioning the class labels 0 $\sim$ 9, and each column shows the results of changing the noise vector $\mathbf{z}$. It can be seen that $G_{CG}$ and $G_{MI}$ produce realistic images of digits and that various image shapes are generated according to $\mathbf{z}$. Images conditioned on 9 in the second and fifth rows look like 7. This can be interpreted as a phenomenon in which the hidden variables for 9 and 7 are placed in adjacent areas on the manifold of the hidden space. 

\begin{table*}[t!]
\centering
\caption{Ablation study of HexaGAN (F1-score)}
\label{tab:ablation}
{\resizebox{0.87\linewidth}{!}
{\begin{tabular}{l|c|c|c|c}

\hline
	\toprule
   \multicolumn{1}{c|}{Method} & Breast & Credit & Wine & Madelon\\
   \midrule
    MLP (HexaGAN w/o $G_{MI} ~\&~ G_{CG} ~\&~ D_{MI_{d+1}}$) & 0.9171 $\pm$ 0.0101 & 0.3404 $\pm$ 0.0080 & 0.9368 $\pm$ 0.0040 & 0.6619 $\pm$ 0.0017 \\
    HexaGAN w/o $G_{CG} ~\&~ D_{MI_{d+1}}$ & 0.9725 $\pm$ 0.0042 & 0.4312 $\pm$ 0.0028 & 0.9724 $\pm$ 0.0065 & 0.6676 $\pm$ 0.0038 \\
    HexaGAN w/o $G_{CG}$ & 0.9729 $\pm$ 0.0007 & 0.4382 $\pm$ 0.0075 & 0.9738 $\pm$ 0.0135 & 0.6695 $\pm$ 0.0043 \\
    HexaGAN w/o $D_{MI_{d+1}}$ & 0.9750 $\pm$ 0.0030 & 0.4604 $\pm$ 0.0097 & 0.9770 $\pm$ 0.0037 & 0.6699 $\pm$ 0.0022 \\
    \textbf{HexaGAN} & \textbf{0.9762} $\pm$ \textbf{0.0021} & \textbf{0.4627} $\pm$ \textbf{0.0040} & \textbf{0.9814} $\pm$ \textbf{0.0059} & \textbf{0.6716} $\pm$ \textbf{0.0019} \\
    \bottomrule
\end{tabular}}
}
\end{table*}

\begin{table*}[t!]
\centering
\caption{Classification performance (F1-score) comparison with other combinations of state-of-the-art methods}
\label{tab:comparison}
{\resizebox{0.75\linewidth}{!}
{\begin{tabular}{l|c|c|c|c}

\hline
	\toprule
   \multicolumn{1}{c|}{Method} & Breast & Credit & Wine & Madelon\\
   \midrule
    MICE + CS + TripleGAN & 0.9417 $\pm$ 0.0044 & 0.3836 $\pm$ 0.0052 & 0.9704 $\pm$ 0.0043 & 0.6681 $\pm$ 0.0028 \\
    GAIN + CS +  TripleGAN & 0.9684 $\pm$ 0.0102 & 0.4076 $\pm$ 0.0038 & 0.9727 $\pm$ 0.0046 & 0.6690 $\pm$ 0.0027 \\
    MICE + SMOTE + TripleGAN & 0.9434 $\pm$ 0.0060 & 0.4163 $\pm$ 0.0029 & 0.9756 $\pm$ 0.0037 & 0.6712 $\pm$ 0.0008 \\
    GAIN + SMOTE + TripleGAN & 0.9672 $\pm$ 0.0063 & 0.4401 $\pm$ 0.0031 & 0.9735 $\pm$ 0.0063 & 0.6703 $\pm$ 0.0032  \\
    \textbf{HexaGAN} & \textbf{0.9762} $\pm$ \textbf{0.0021} & \textbf{0.4627} $\pm$ \textbf{0.0040} & \textbf{0.9814} $\pm$ \textbf{0.0059} & \textbf{0.6716} $\pm$ \textbf{0.0019} \\
    \bottomrule
\end{tabular}}
}
\end{table*}

\subsection{Classification performance}
HexaGAN works without any problem for multi-class classification, but for the convenience of the report, we tested only binary classifications. The breast and credit datasets are imbalanced with a large number of negative samples. The wine dataset has three classes, and it was tested by binarizing the label 1 as negative, and labels 2 and 3 as positive to calculate an F1-score. The wine dataset was imbalanced with a large number of positive samples. Madelon is a balanced synthetic dataset that randomly assigns binary labels to 32 clusters on 32 vertices of a 5-dimensional hypercube.

\subsubsection{Ablation study}
The components affecting the classification performance of HexaGAN are $G_{MI}$ to fill in missing data, $G_{CG}$ to perform conditional generation, and $D_{MI}(\cdot)_{d+1}$ to enable semi-supervised learning. Table \ref{tab:ablation} compares the classification performance depending on the removal of these components. In the case of MLP, which is equivalent to HexaGAN without any of these three components, missing data were filled in with values sampled uniformly from [0,1]. 

As a result, MLP shows the worst performance. When HexaGAN contains $G_{CG}$ (from the second row to the fourth row), the biggest performance improvement is shown in the credit data which is the most imbalanced. The more the components included in HexaGAN, the higher the classification performance obtained. HexaGAN with all components shows the highest performance on every dataset. Our delicately devised architecture improves the classification performance by up to 36\%. It offers the advantage that any classifier that is state-of-the-art in a controlled environment can be plugged into the proposed framework, and the classifier will perform at its highest capacity.

\begin{figure}[t]\centering
\includegraphics[width=0.8\linewidth]{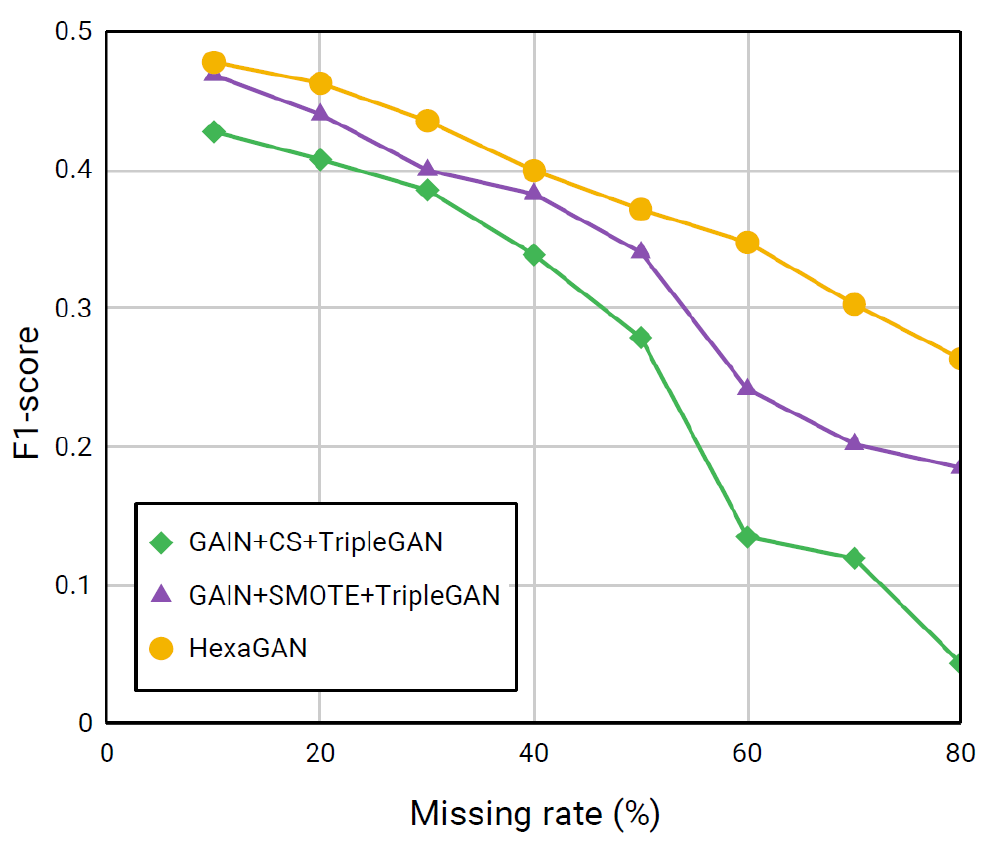} 
\caption{Classification performance (F1-score) comparison with respect to the missing rate with the credit dataset}
\label{fig:rate_f1}
\vspace{-1em}
\end{figure}

\subsubsection{Comparison with other combinations}
In this experiment, we compared the classification performance of HexaGAN with those of combinations of state-of-the-art methods for the three problems. For missing data imputation, we used MICE, which showed the best performance among machine learning based methods, and GAIN, which showed the best performance among deep generative models. For class imbalance, we used the cost sensitive loss (CS) and oversampled the minority class in a batch using SMOTE. We adopted the TripleGAN for semi-supervised learning. The classifier of TripleGAN used the same architecture as $C$ of HexaGAN for a fair comparison. 

As shown in Table \ref{tab:comparison}, HexaGAN shows significantly better performance than the combinations of existing methods in cascading form (up to a 5\% improvement). In addition, the madelon dataset is balanced; thus, comparing HexaGAN without $G_{CG}$ (the third row of Table \ref{tab:ablation}) with the combination of MICE, CS, and TripleGAN (the first row of Table \ref{tab:comparison}) and the combination of GAIN, CS, and TripleGAN (the second row of Table \ref{tab:comparison}) shows the classification performance with respect to imputation methods. We confirm that the imputation method of HexaGAN guarantees better classification performance than the other imputation methods. 

\subsubsection{Classification performance with respect to missing rate}
Figure \ref{fig:rate_f1} compares the classification performance of HexaGAN with those of competitive combinations for various missing rates in the credit dataset. We used the combination of GAIN, CS, and TripleGAN and the combination of GAIN, SMOTE, and TripleGAN as benchmarks. According to the results, HexaGAN outperforms the benchmarks for all missing rates. Moreover, our method shows a larger performance gap compared to the benchmarks for high missing rates. This means that HexaGAN works robustly in situations in which only little information is available.

%% file: 5_conclusion.tex
To interactively overcome the three main problems in real world classification (missing data, class imbalance, and missing label), we define the three problems from the perspective of missing information. Then, we propose a HexaGAN framework wherein six neural networks are actively correlated with others, and design several loss functions that maximize the utilization of any incomplete data. Our proposed method encourages more powerful performance in both imputation and classification than existing state-of-the-art methods. Moreover, HexaGAN is a one-stop solution that automatically solves the three problems commonly presented in real world classification. For future work, we plan to extend HexaGAN to time series datasets such as electronic health records.

%% file: appendix.tex
\section{Proofs}
\subsection{Global optimality of $p(\mathbf{x}|\mathbf{m}_i=1)=p(\mathbf{x}|\mathbf{m}_i=0)$ for HexaGAN}

\textbf{Proof of Theorem 1:} Let $D_{MI}(\cdot)$ be $D(\cdot)$, and $G_{MI}(E(\cdot))$ be $G(\cdot)$ for convenience.

The min-max loss of HexaGAN for missing data imputation is given by:

\begin{align}
    V_{MI}(D,G) &= \mathbb{E}_{\mathbf{x},\mathbf{z},\mathbf{m}}\left[\mathbf{m}^TD(G(\tilde{\mathbf{x}}|\mathbf{m}))-(\mathbf{1-m})^TD(G(\tilde{\mathbf{x}}|\mathbf{m}))\right] \\
    &= \mathbb{E}_{\hat{\mathbf{x}},\mathbf{m}}\left[\mathbf{m}^TD(\hat{\mathbf{x}})-(\mathbf{1-m})^TD(\hat{\mathbf{x}})\right] \\
    &= \int_\mathcal{\hat{X}}\sum_{\mathbf{m}\in \lbrace0,1\rbrace^d}\left(\mathbf{m}^TD(\mathbf{x})-(1-\mathbf{m})^TD(\mathbf{x})\right)p(\mathbf{x}|\mathbf{m}) d\mathbf{x} \\
    &= \int_\mathcal{\hat{X}}\sum_{\mathbf{m}\in \lbrace0,1\rbrace^d}\left(\sum_{i : m_i=1}D(\mathbf{x})_i-\sum_{i : m_i=0}D(\mathbf{x})_i\right)p(\mathbf{x}|\mathbf{m}) d\mathbf{x} \\
    &= \int_\mathcal{\hat{X}}\sum_{i=1}^d\left(D(\mathbf{x})_i\sum_{\mathbf{m}:m_i=1 }p(\mathbf{x}|\mathbf{m})-D(\mathbf{x})_i\sum_{\mathbf{m}:m_i=0}p(\mathbf{x}|\mathbf{m})\right) d\mathbf{x} \\
    &= \int_\mathcal{\hat{X}}\sum_{i=1}^d D(\mathbf{x})_i p(\mathbf{x}|m_i=1)-D(\mathbf{x})_i p(\mathbf{x}|m_i=0) d\mathbf{x} \label{eq6} \\
    &= \int_\mathcal{\hat{X}}\sum_{i=1}^d\left(p(\mathbf{x}|m_i=1)-p(\mathbf{x}|m_i=0)\right)D(\mathbf{x})_i d\mathbf{x} \label{eq7}
\end{align}
For a fixed G, the optimal discriminator $D(\mathbf{x})_i$ which maximizes $V_{MI}(D,G)$ is such that:
\begin{align}
    D^*_G(\mathbf{x})_i =
    \begin{cases}
        1, & \text{if } p(\mathbf{x}|m_i=1) \geq p(\mathbf{x}|m_i=0)\\ 
        0, & \text{otherwise}
    \end{cases}
\end{align}
Plugging $D^*_G$ back into Equation~\ref{eq7}, we get:
\begin{align}
    V_{MI}(D^*_G,G) &= \int_\mathcal{\hat{X}}\sum_{i=1}^d\left(p(\mathbf{x}|m_i=1)-p(\mathbf{x}|m_i=0)\right)D^*_G(\mathbf{x})_i d\mathbf{x} \\
    &= \sum_{i=1}^d\int_{\{\mathbf{x}|p(\mathbf{x}|m_i=1) \geq p(\mathbf{x}|m_i=0)\}}\left(p(\mathbf{x}|m_i=1)-p(\mathbf{x}|m_i=0)\right)d\mathbf{x} \label{eq11}
\end{align}
Let $\mathcal{X}=\{\mathbf{x}|p(\mathbf{x}|m_i=1) \geq p(\mathbf{x}|m_i=0)\}$. To minimize Equation~\ref{eq11}, we need to set $p(\mathbf{x}|m_i=1) = p(\mathbf{x}|m_i=0)$ for $\mathbf{x}\in\mathcal{X}$. 

Then, when we consider $\mathcal{X}^c$, the complement of $\mathcal{X}$, $p(\mathbf{x}|m_i=1) < p(\mathbf{x}|m_i=0)$ for $\mathbf{x}\in\mathcal{X}^c$. Since both probability density functions should integrate to 1,
\begin{align}
    \int_{\mathcal{X}^c}p(\mathbf{x}|m_i=1)d\mathbf{x}=\int_{\mathcal{X}^c}p(\mathbf{x}|m_i=0)d\mathbf{x}
\end{align}
However, this is a contradiction, unless $\lambda({X}^c)=0$ where $\lambda$ is the Lebesgue measure. This finishes the proof. \hfill $\square$

\subsection{Optimization of components for imputation}
From Equation~\ref{eq6},
\begin{align}
    V_{MI}(D,G)_i &= \int_\mathcal{\hat{X}}p(\mathbf{x}|m_i=1) D(\mathbf{x})_i - p(\mathbf{x}|m_i=0) D(\mathbf{x})_i d\mathbf{x} \\
    &= \mathbb{E}_{\tilde{\mathbf{x}},\mathbf{z},\mathbf{m}}\left[m_i \cdot D(G(\tilde{\mathbf{x}}|\mathbf{m}))_i\right]-\mathbb{E}_{\tilde{\mathbf{x}},\mathbf{z},\mathbf{m}}\left[(1-m_i) \cdot D(G(\tilde{\mathbf{x}}|\mathbf{m}))_i\right]
\end{align}
G is then trained according to $\min_G \sum_{i=1}^d V_{MI}(D,G)_i$, and D is trained according to $\max_D \sum_{i=1}^d V_{MI}(D,G)_i$.

\subsection{Relation between pseudo-labeling and the ODM cost}
\textbf{Proof of Theorem 2:} Optimizing the adversarial loss functions $L_C$ and $L_{D_{MI}}^{d+1}$ are equivalent to minimizing the Earth Mover distance between $\mathrm{Distr}[C(\mathbf{\hat{x}}_u)]$ and $\mathrm{Distr}[\mathbf{y}]$, where $\mathrm{Distr[\mathbf{\cdot}]}$ denotes the distribution of a random variable. 

Since converging the Earth Mover distance $W(p,q)$ to zero implies that the two distributions $p$ and $q$ are equal, the following proposition holds
\begin{align}
W(\mathrm{Distr}[C(\mathbf{\hat{x}}_u)],\mathrm{Distr}[\mathbf{y}]) \rightarrow 0 ~~\Rightarrow~~ \mathrm{Distr}[C(\mathbf{\hat{x}}_u)] = \mathrm{Distr}[\mathbf{y}]
\end{align}
This means that minimizing the Earth Mover distance $W(\mathrm{Distr}[C(\mathbf{\hat{x}}_u)],\mathrm{Distr}[\mathbf{y}])$ matches the distributions of the outputs. Therefore, the adversarial losses of $D_{MI}$ and $C$ satisfy the definition of the output distribution matching (ODM) cost function, concluding the proof. \hfill $\square$

\section{Training of HexaGAN in details}
\subsection{Dataset description}
Table \ref{tab:dataset} presents the dataset descriptions used in the experiments. The imbalance ratio of the wine dataset is calculated from the binarized classes by combining classes 2 and 3 into one class, and the numbers of data in the three classes are 59, 71, and 48, respectively.

\begin{table*}[h!]
\centering
\caption{Dataset description. The imbalance ratio indicates the ratio of the number of instances in the majority class to the number of instances in the minority class.}
\label{tab:dataset}
{\resizebox{0.65\textwidth}{!}
{\begin{tabular}{c|c|c|c}
\hline
	\toprule
	Dataset & \# of features & \# of instances & Imbalance ratio (1:$x$) \tabularnewline \midrule
	Breast & 30 & 569 & 1.68 \tabularnewline
	Credit & 23 & 30,000 & 3.52 \tabularnewline
	Wine (with binarized class) & 13 & 178 & 2.02 \tabularnewline
	Madelon & 500 & 4,400 & 1.00 \tabularnewline
	\bottomrule
\end{tabular}}
}
\end{table*}

\subsection{Training procedure}
Each component of the whole system is updated in order. We should note that the distribution of $\mathbf{h}_l$ is altered by the updating of E; thus, we updated $G_{CD}$ and $D_{CG}$ several times when the other components are updated once, as shown in Algorithm \ref{alg:procedure}. We set the number of iterations for the conditional generation per an iteration for the other components to 10 and the number of iterations for discriminators per an iteration for generators to 5 in our experiments.

\begin{algorithm}[h!]
    \SetKwInOut{Input}{Require}
    \Input{$n_{CG}$ - the number of iterations for the conditional generation per an iteration for the other components;
    \newline $n_{critic}$ - the number of iterations for discriminators per an iteration for generators}
    \caption{Training procedure of HexaGAN}
    \label{alg:procedure}
\begin{algorithmic}
   \WHILE{training loss is not converged}
   \STATE{\textbf{(1) Missing data imputation}}
   \FOR{$k = 1, ..., n_{critic}$}
   \STATE Update $D_{MI}$ using stochastic gradient descent (SGD)
   \STATE $\nabla_{D_{MI}} \mathcal{L}_{D_{MI}} + \mathcal{L}_{D_{MI}}^{d+1} + \lambda_1 \mathcal{L}_{\mathrm{GP}_{MI}}$
   \ENDFOR
   \STATE Update $E$ using SGD
   \STATE $\nabla_{E} \mathcal{L}_{G_{MI}} + \alpha_1 \mathcal{L}_{\mathrm{recon}}$
   \STATE Update $G_{MI}$ using SGD 
   \STATE $\nabla_{G_{MI}} \mathcal{L}_{G_{MI}} + \alpha_1 \mathcal{L}_{\mathrm{recon}}$
   \newline
   \STATE{\textbf{(2) Conditional generation}}
   \FOR{$i = 1, ..., n_{CG}$}
   \FOR{$j = 1, ..., n_{critic}$}
   \STATE Update $D_{CG}$ using SGD
   \STATE $\nabla_{D_{CG}} \mathcal{L}_{D_{CG}} + \lambda_2 \mathcal{L}_{\mathrm{GP}_{CG}}$
   \ENDFOR
   \STATE Update $G_{CG}$ using SGD
   \STATE $\nabla_{G_{CG}} \mathcal{L}_{G_{CG}} + \alpha_2 \mathcal{L}_{G_{MI}} + \alpha_3 \mathcal{L}_{\mathrm{CE}}(\mathbf{\hat{x}}_{c}, \mathbf{y}_{c})$
   \ENDFOR
   \newline
   \STATE{\textbf{(3) Semi-supervised classification}}
   \STATE Update $C$ using SGD 
   \STATE $\nabla_{C} \mathcal{L}_{\mathrm{CE}}(\mathbf{\hat{x}}_{l,c}, \mathbf{y}_{l,c}) + \alpha_4 \mathcal{L}_C$
   \ENDWHILE
\end{algorithmic}
\end{algorithm}

\subsection{Architecture of HexaGAN}
Excluding the experiments in Sections 4.1.2 and 4.2, all six components used an architecture with three fully-connected layers. The number of hidden units in each layer is $d$, $d/2$, and $d$. As an activation function, we use the rectified linear unit (ReLU) function for all hidden layers and the output layer of $E$ and $G_{CG}$, the sigmoid function for the output layer of $G_{MI}$ and $D_{CG}$, no activation function for the output layer of $D_{MI}$, and the softmax function for the output layer of $C$.

Table \ref{tab:architecture} describes the network architectures used in Sections 4.1.2 and 4.2. In the table, FC($n$) denotes a fully-connected layer with $n$ output units. Conv($n$, $k \times k$, $s$) denotes a convolutional network with $n$ feature maps, filter size $k \times k$, and stride $s$. Deconv($n$, $k \times k$, $s$) denotes a deconvolutional network with $n$ feature maps, filter size $k \times k$, and stride $s$.

\begin{table*}[h]
\centering
\caption{Convolutional neural network architectures used for the MNIST dataset}
\label{tab:architecture}
{\begin{tabular*}{0.93\textwidth}{c|c|c|c|c|c}
\hline
   \toprule
    $G_{CG}$ & $D_{CG}$ & $E$ & $G_{MI}$ & $D_{MI}$ & $C$ \\
   \midrule
    FC(512) & FC(1024) & Conv(32, 5$\times$5, 2) & Deconv(64, 5$\times$5, 2) & Conv(32, 5$\times$5, 2) & Conv(32, 5$\times$5, 2) \\
    ReLU & ReLU & ReLU & ReLU & ReLU & ReLU \\
    FC(1024) & FC(512) & Conv(64, 5$\times$5, 2) & Deconv(32, 5$\times$5, 2) & Conv(64, 5$\times$5, 2) & Conv(64, 5$\times$5, 2) \\
    ReLU & ReLU & ReLU & ReLU & ReLU & ReLU \\
    FC(2048) & FC(1) & Conv(128, 5$\times$5, 2) & Deconv(1, 5$\times$5, 2) & Conv(128, 5$\times$5, 2) & Conv(128, 5$\times$5, 2) \\
    ReLU & Sigmoid & ReLU & ReLU & ReLU & ReLU \\
     &  &  & FC(784) & FC(785) & FC(10) \\
     &  &  & Sigmoid & Sigmoid & Softmax \\
    \bottomrule
\end{tabular*}}
\end{table*}

\section{Additional experiments}
\subsection{Learning curve analysis on missing data imputation}
\begin{figure}[h!]
\centering
\subfigure[Comparison of the gradient penalty\label{fig:gp}]{\includegraphics[width=0.49\linewidth]{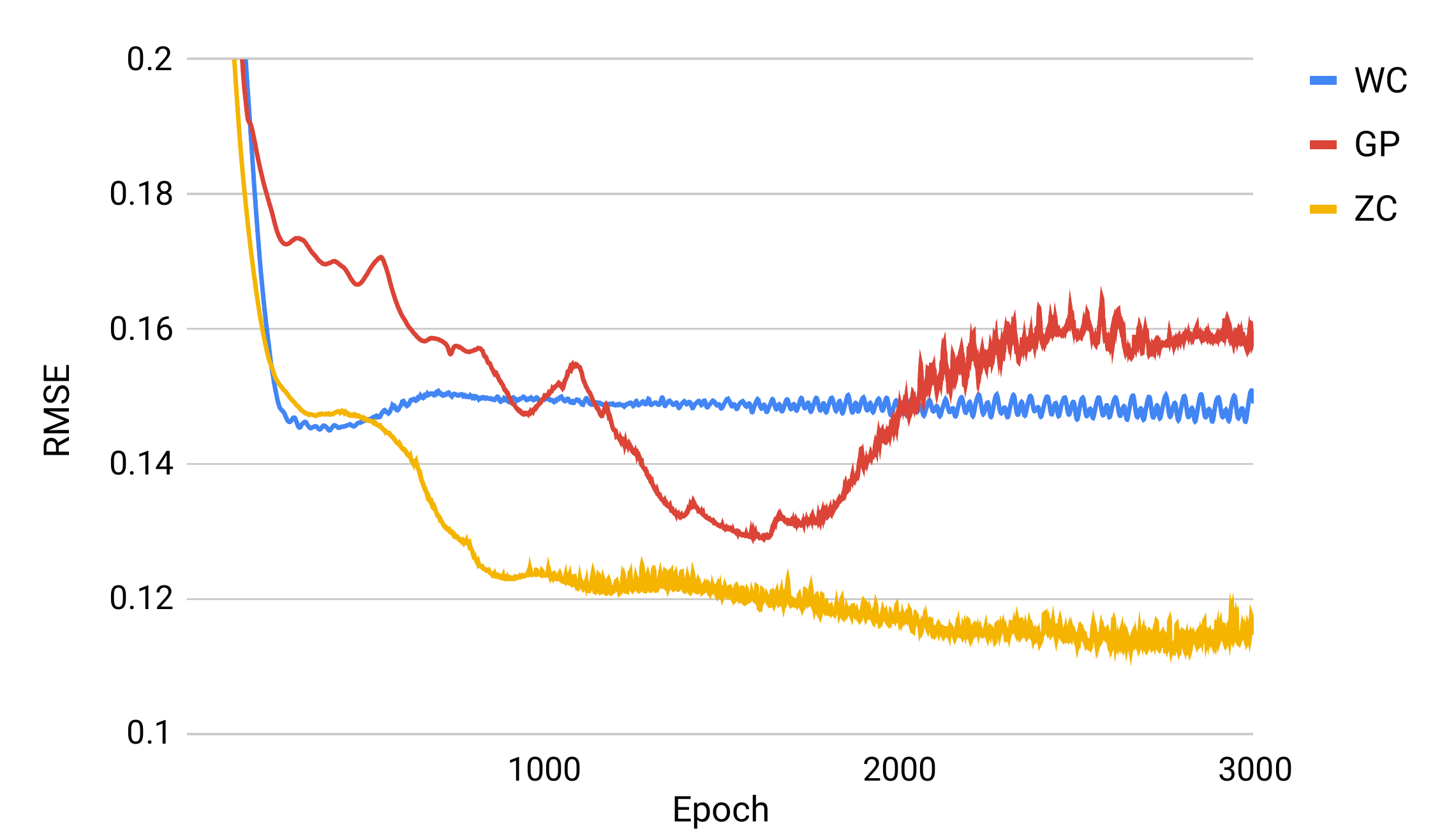}}
\subfigure[Comparison of the adversarial loss and optimizer\label{fig:adv_loss}]{\includegraphics[width=0.49\linewidth]{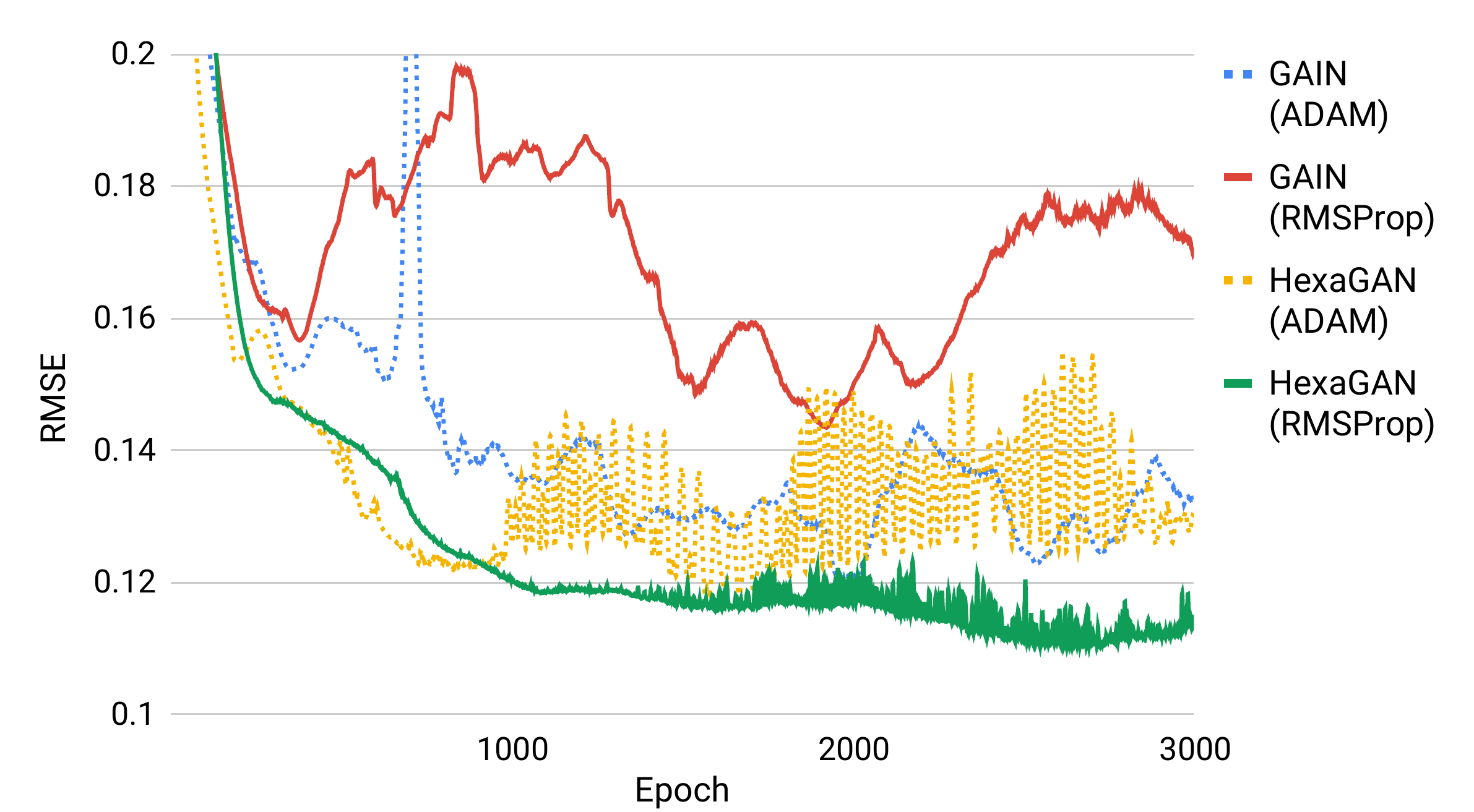}}
\caption{Learning curve comparison for the optimal GAN imputation method}
\label{fig:imputation_select}
\end{figure}

Using the breast dataset, we measured the RMSE to evaluate the imputation performance of the proposed adversarial losses ($\mathcal{L}_{D_{MI}}$, $\mathcal{L}_{G_{MI}}$). We excluded $\mathcal{L}_{\mathrm{recon}}$ from the losses of $E$ and $G_{MI}$ and compared the learning curves of weight clipping (WC) proposed by \citetapndx{wgan}, the modified gradient penalty (GP) of \citetapndx{wgan-gp}, and the modified zero-centered gradient penalty (ZC, ours) to determine the most appropriate gradient penalty for our framework. As shown in Figure \ref{fig:gp}, ZC shows stable and good performance (small RMSE). In Figure \ref{fig:adv_loss}, we plot learning curves to accurately compare the adversarial losses of GAIN and HexaGAN. We also compare the two optimizers ADAM \citeapndx{adam} and RMSProp \citeapndx{rmsprop}. Our experiment shows that RMSProp is a more stable optimizer than ADAM, and HexaGAN produces a more stable and better imputation performance than GAIN.

\subsection{Imputation performance with respect to the missing rate}
\begin{figure}[h!]\centering
\includegraphics[width=0.45\linewidth]{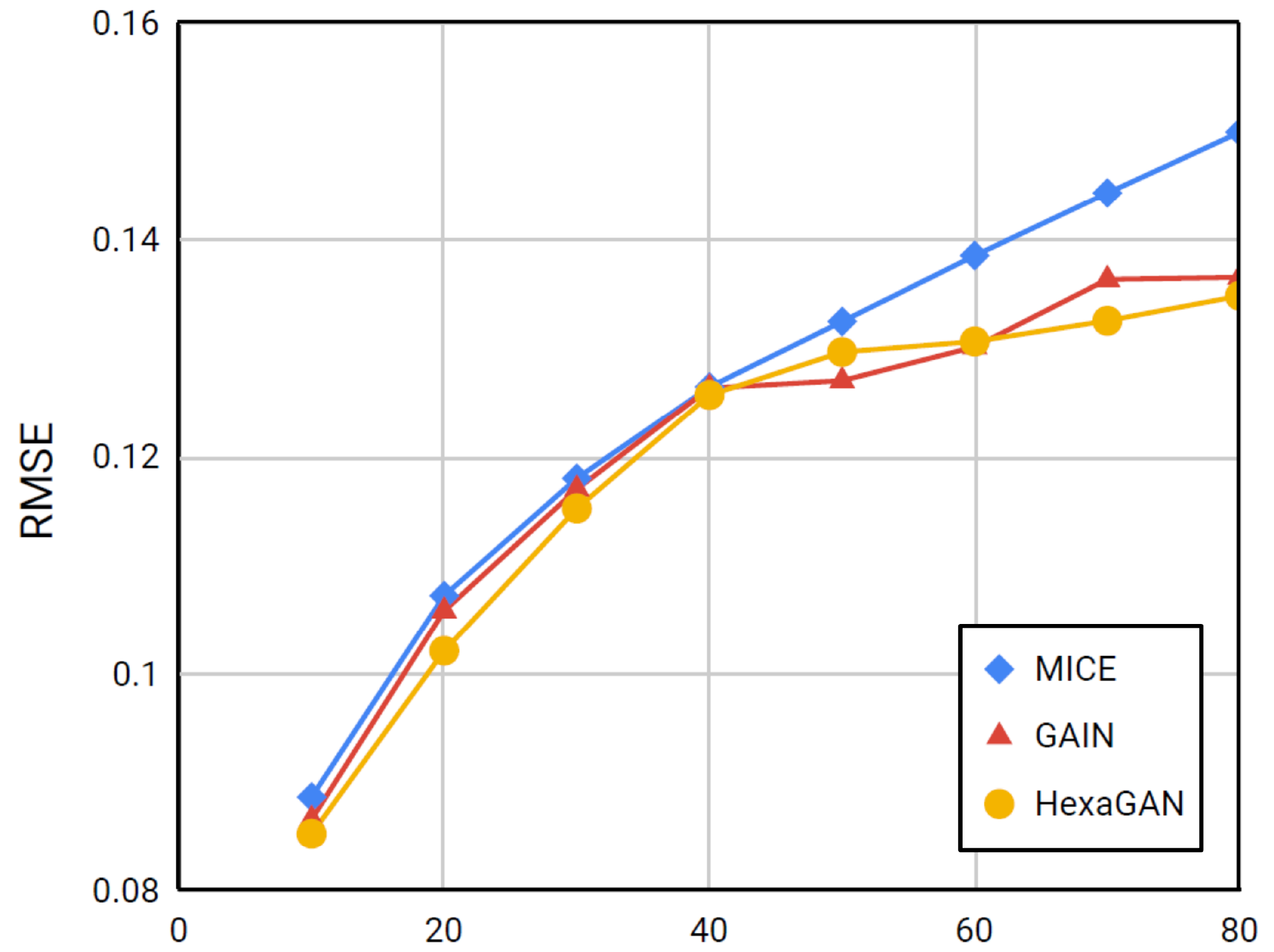} 
\caption{Imputation performance (RMSE) comparison with respect to the missing rate with the credit dataset}
\label{fig:rate_rmse}
\end{figure}

We measured the imputation performance of HexaGAN for various missing rates in the credit dataset. To compare the performance with those of competitive benchmarks, we used MICE, which is a state-of-the-art machine learning algorithm, and GAIN, which is a state-of-the-art deep generative model. As seen in Figure \ref{fig:rate_rmse}, HexaGAN shows the best performance for all missing rates except 50\%. The comparison of MICE and HexaGAN shows that the gap between the performances of the two methods increases at higher missing rates; therefore, HexaGAN is more robust when there is less information available.

\subsection{tSNE analysis on conditional generation}
\begin{figure*}[h!]
\includegraphics[width=1\linewidth]{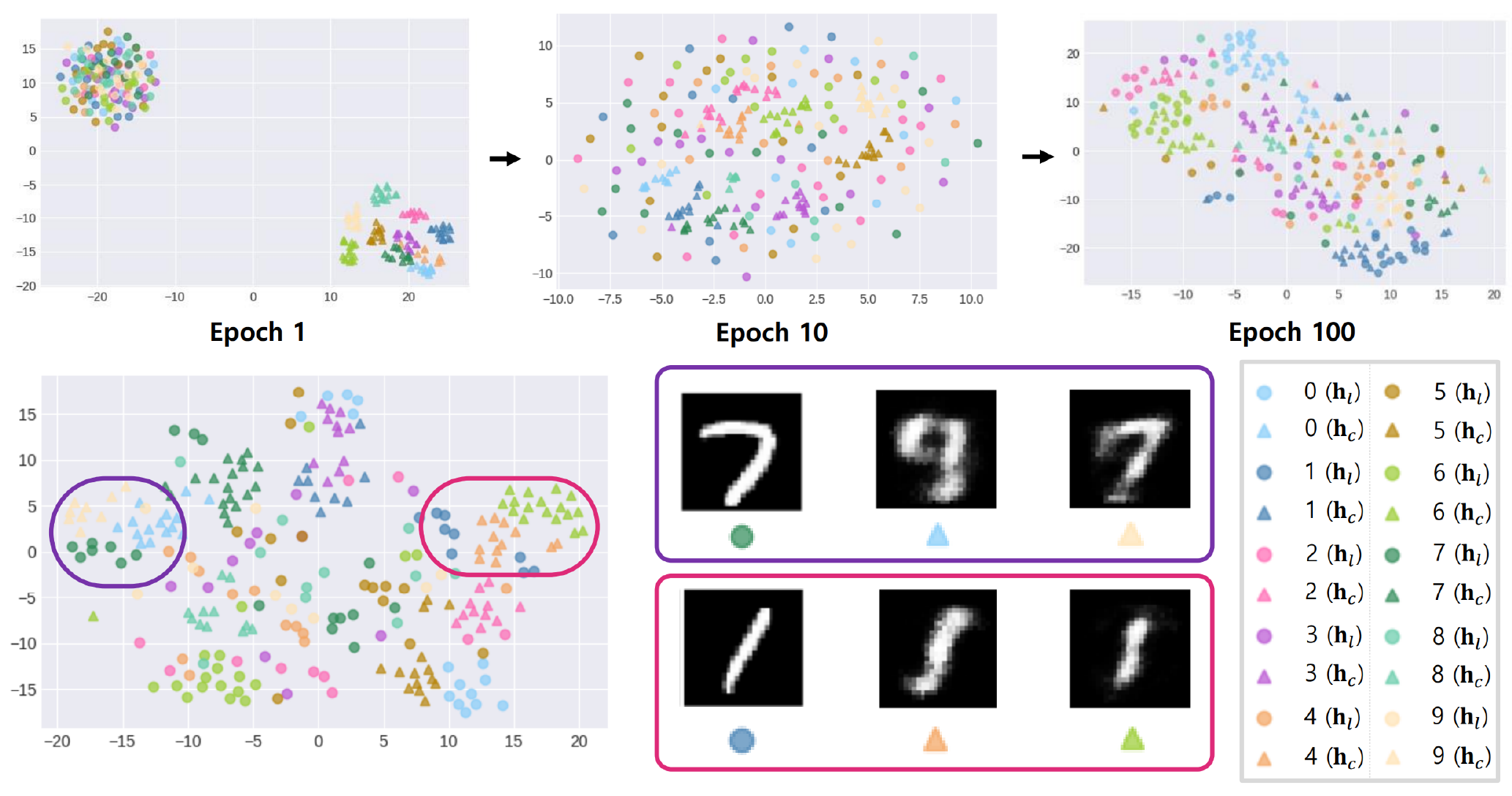} 
\caption{tSNE analysis with the MNIST dataset}
\label{fig:tsne}
\end{figure*}

Figure \ref{fig:tsne} is the complete version of the tSNE analysis in Section 4.2.1. The tSNE plot below shows an analysis of the manifold of the hidden space. We confirm that the synthetic data around the original data looks similar to the original data. Therefore, it can be seen that $E$ learns the data manifold well in the hidden space.

\subsection{Sensitivity analysis of loss functions}
We performed diverse experiments by tuning the hyperparameter of each loss term for the missing data imputation and conditional generation experiments. We utilized the credit dataset and measured the RMSE and F1-score. The first two rows of Table \ref{tab:sensitivity} show the imputation performances (RMSE) acheived by tuning hyperparameters $\alpha_1$ and $\lambda_1$, which are multiplied by the auxiliary loss terms for missing data imputation ($\mathcal{L}_{\mathrm{recon}}$ and $\mathcal{L}_{\mathrm{GP}_{MI}}$, respectively). The results show that HexaGAN achieves the best missing data imputation performance when both $\alpha_1$ and $\lambda_1$ are set to 10. The last two rows of Table \ref{tab:sensitivity} present the classification performances (F-score) acheived by tuning hyperparameters $\alpha_2$ and $\alpha_3$, which are multiplied by the auxiliary losses for conditional generation ($\mathcal{L}_{G_{MI}}$ and $\mathcal{L}_{\mathrm{CE}}(\mathbf{\hat{x}}_{c}, \mathbf{y}_{c})$, respectively). As a result, the best classification performance is obtained when $\alpha_2$ and $\alpha_3$ are the default values in our paper, at 1 and 0.01, respectively.

\begin{table*}[h!]
\centering
\caption{Sensitivity analysis of the loss functions with the credit dataset}
\label{tab:sensitivity}
{\resizebox{0.65\textwidth}{!}
{\begin{tabular}{c|c|cccc}

\hline
	\toprule
	Hyperparameter (Loss) & Setting & 1 & 2 & 3 & 4 \tabularnewline \midrule
	$\alpha_1$ $\left(\mathcal{L}_{\mathrm{recon}}\right)$ & Value & 0 & 1 & \textbf{10} & 100 \tabularnewline
	 & RMSE & 0.1974 & 0.1108 & \textbf{0.1022} & 0.1079 \tabularnewline
	\midrule
	$\lambda_1$ $\left(\mathcal{L}_{\mathrm{GP}_{MI}}\right)$ & Value & 0 & 1 & \textbf{10} & 100 \tabularnewline
	 & RMSE & 0.1110 & 0.1097 & \textbf{0.1022} & 0.1081 \tabularnewline
	\midrule \midrule
	$\alpha_2$ $\left(\mathcal{L}_{G_{MI}}\right)$ & Value & 0 & \textbf{1} & 10 & 100 \tabularnewline
	 & F1-score & 0.4535 & \textbf{0.4627} & 0.4585 & 0.4523 \tabularnewline
	\midrule
	$\alpha_3$ $\left(\mathcal{L}_{\mathrm{CE}}(\mathbf{\hat{x}}_{c}, \mathbf{y}_{c})\right)$ & Value & 0 & \textbf{0.01} & 0.1 & 1 \tabularnewline
	 & F1-score & 0.4535 & \textbf{0.4627} & 0.4585 & 0.4523 \tabularnewline
    \bottomrule
\end{tabular}}
}
\end{table*}

\subsection{Statistical significance}
We conducted statistical tests for Tables 1, 2, and 3 in the original paper. Because the results of the experiment could not meet the conditions of normality and homogeneity of variance tests, we used a non-parametric test, the Wilcoxon rank sum test. We additionally measured the effect size using Cohen's d. We validated that all the experiments are statistically significant or showed large or medium effect size, except for GAIN vs. HexaGAN for the wine dataset in Table 1, HexaGAN without $D_{MI}$ vs. HexaGAN for the breast and credit datasets in Table 2, and MICE + SMOTE + TripleGAN vs. HexaGAN for the madelon dataset in Table 3.

\subsection{Classification performance with the CelebA dataset}
We used a more challenging dataset, CelebA. It is a high-resolution face dataset for which it is more difficult to impute missing data. CelebA consists of 40 binary attributes with various imbalance ratios (1:1 $\sim$ 1:43). We used 50,000 and 10,000 labeled and unlabeled training images, respectively, and 10,000 test images. The size of each image is 218x178x3, which means that the data dimension is 116,412. Therefore, we could evaluate our method on the setting where the data dimension is less than the sample size. Then, half of the elements were removed from each image under the 50\% missingness (MCAR) assumption. 

For comparison, we utilized a class rectification loss (CRL) \citeapndx{crl} which is the most recent method developed for the class imbalance problem. Since an image has 40 labels simultaneously, we simply balanced the class of data entered into $C$ by setting the class condition to $\mathbf{1}-\mathbf{y}$. Additionally, the data dimension was too large to calculate $\mathcal{L}_{\mathrm{GP}_{MI}}$, therefore we replaced the regularization for discriminator learning with weight clipping. We measured the F1-scores for 40 attributes for three cases: GAIN + TripleGAN, GAIN + CRL + TripleGAN, and HexaGAN. The same structure and hyperparameters were used for the classifier for a fair comparison. Table \ref{tab:celeba} shows the imbalance ratio of each attribute and the classification performance (F1-score) of each combination. Comparing the average F1-score of 40 attributes, GAIN + TripleGAN shows a performance of 0.5152, GAIN + CRL + TripleGAN has a performance of 0.5519, and HexaGAN has a performance of 0.5826. HexaGAN outperforms all the compared methods.


\begin{table*}[h]
\centering
\caption{Classification performance comparison with the CelebA dataset (F1-score)}
\label{tab:celeba}
{\resizebox{0.9\textwidth}{!}
{\begin{tabular}{c|c|c|c|c}

\hline
	\toprule
	Attribute & Imb. ratio (1:$x$) & GAIN + TripleGAN & GAIN + CRL + TripleGAN & ~~~~HexaGAN~~~~ \tabularnewline \midrule
	Arched eyebrows & 3 & 0.53 & 0.50 & \textbf{0.55} \tabularnewline
    Attractive & 1 & \textbf{0.78} & 0.74 & 0.74 \tabularnewline
    Bags under eyes & 4 & 0.30 & 0.44 & \textbf{0.49} \tabularnewline
    Bald & 43 & 0.37 & \textbf{0.42} & 0.35 \tabularnewline
    Bangs & 6 & 0.70 & \textbf{0.77} & 0.71 \tabularnewline
    Big lips & 3 & 0.17 & 0.20 & \textbf{0.39} \tabularnewline
    Big nose & 3 & 0.41 & 0.47 & \textbf{0.49} \tabularnewline
    Black hair & 3 & 0.67 & \textbf{0.72} & 0.69 \tabularnewline
    Blond hair & 6 & \textbf{0.77} & 0.74 & 0.71 \tabularnewline
    Blurry & 18 & 0.02 & \textbf{0.16} & 0.15 \tabularnewline
    Brown hair & 4 & 0.49 & 0.49 & \textbf{0.57} \tabularnewline
    Bushy eyebrows & 6 & 0.48 & \textbf{0.55} & 0.49 \tabularnewline
    Chubby & 16 & \textbf{0.49} & 0.33 & 0.45 \tabularnewline
    Double chin & 20 & 0.34 & 0.36 & \textbf{0.46} \tabularnewline
    Eyeglasses & 14 & 0.64 & \textbf{0.81} & 0.79 \tabularnewline
    Goatee & 15 & 0.41 & 0.48 & \textbf{0.50} \tabularnewline
    Gray hair & 23 & 0.46 & 0.55 & \textbf{0.59} \tabularnewline
    Heavy makeup & 2 & 0.80 & \textbf{0.84} & \textbf{0.84} \tabularnewline
    High cheekbones & 1 & 0.78 & 0.79 & \textbf{0.80} \tabularnewline
    Male & 1 & 0.91 & \textbf{0.93} & \textbf{0.93} \tabularnewline
    Mouth slightly open & 1 & 0.81 & \textbf{0.83} & 0.82 \tabularnewline
    Mustache & 24 & 0.36 & \textbf{0.58} & 0.49 \tabularnewline
    Narrow eyes & 8 & 0.17 & 0.25 & \textbf{0.28} \tabularnewline
    No beard & 5 & 0.95 & \textbf{0.95} & 0.92 \tabularnewline
    Oval face & 3 & 0.16 & 0.24 & \textbf{0.47} \tabularnewline
    Pale skin & 22 & 0.34 & \textbf{0.45} & 0.39 \tabularnewline
    Pointy nose & 3 & 0.49 & 0.31 & \textbf{0.52} \tabularnewline
    Receding hairline & 11 & 0.22 & \textbf{0.46} & 0.44 \tabularnewline
    Rosy cheeks & 14 & 0.45 & 0.53 & \textbf{0.55} \tabularnewline
    Shadow & 8 & 0.45 & \textbf{0.49} & 0.46 \tabularnewline
    Sideburns & 17 & 0.50 & 0.58 & \textbf{0.60} \tabularnewline
    Smiling & 1 & 0.85 & \textbf{0.87} & \textbf{0.87} \tabularnewline
    Straight hair & 4 & 0.30 & 0.07 & \textbf{0.38} \tabularnewline
    Wavy hair & 2 & 0.52 & 0.50 & \textbf{0.57} \tabularnewline
    Wearing earrings & 4 & 0.44 & 0.48 & \textbf{0.53} \tabularnewline
    Wearing hat & 19 & 0.65 & 0.67 & \textbf{0.70} \tabularnewline
    Wearing lipstick & 1 & \textbf{0.88} & \textbf{0.88} & \textbf{0.88} \tabularnewline
    Wearing necklace & 7 & 0.04 & 0.11 & \textbf{0.35} \tabularnewline
    Wearing necktie & 13 & 0.62 & \textbf{0.65} & 0.63 \tabularnewline
    Young & 4 & \textbf{0.89} & \textbf{0.89} & 0.76 \tabularnewline \midrule
    Mean & - & 0.5152 & 0.5519 & \textbf{0.5826} \tabularnewline
    \bottomrule
\end{tabular}}
}
\end{table*}